\documentclass[letterpaper,10pt,conference]{ieeeconf}

\IEEEoverridecommandlockouts
\usepackage{amsmath}
\usepackage{amssymb}
\usepackage{bm}
\usepackage{graphicx}
\usepackage{booktabs}
\usepackage{multirow}
\usepackage{cite}
\usepackage[hidelinks]{hyperref}
\hypersetup{
  pdftitle={Toward Physically Grounded JEPA World Models for Goal-Conditioned Robotic Planning},
  pdfauthor={Muyuan Liu, Yue Huang, Zheng Liang, Xiang Gao},
  pdfsubject={},
  pdfkeywords={}
}

\title{\LARGE \bf
Toward Physically Grounded JEPA World Models for Goal-Conditioned Robotic Planning
}

\author{Muyuan Liu, Yue Huang, Zheng Liang, Xiang Gao%
\thanks{Muyuan Liu and Yue Huang contributed equally to this work.
All authors are with Genisom AI, Beijing, China.
Corresponding author: Xiang Gao
\mbox{(e-mail: {\tt\small gao.xiang.thu@gmail.com})}.}%
}

\begin{document}

\maketitle
\thispagestyle{empty}
\pagestyle{empty}

\begin{abstract}
Action-conditioned JEPA world models enable planning toward visually specified
goals without reconstructing future pixels, yet latent prediction alone does not
explicitly encourage the learned representations to retain information relevant
to robotic control. We introduce an end-to-end JEPA world model that augments latent prediction with inverse dynamics (IDM) and state alignment (SA).
While inverse dynamics discourages latent collapse and makes latent transitions informative of the actions
that produced them, state alignment grounds consecutive
representations in their associated physical configuration and motion. Across
four benchmark tasks, our model attains the highest success rates on TwoRoom
(100\%), PushT (98\%), and OGBench-Cube (87\%), while performing comparably to
LeWorldModel on Reacher. Our ablation further shows that adding state alignment
consistently improves planning success over IDM alone across all four tasks.
Although LeWorldModel, our primary
baseline, attains higher average straightening on OGBench-Cube,
transition-subspace analysis shows that its transition
energy is concentrated in a substantially lower-dimensional subspace. Our
state-aligned model exhibits a higher effective transition dimension than
LeWorldModel and improves planning over IDM alone, supporting state alignment as an effective
complement to inverse dynamics for robotic planning. We release our code at https://github.com/zsibot/gawm.

\end{abstract}

\begin{figure*}[t]
  \centering
  \includegraphics[width=0.85\textwidth]{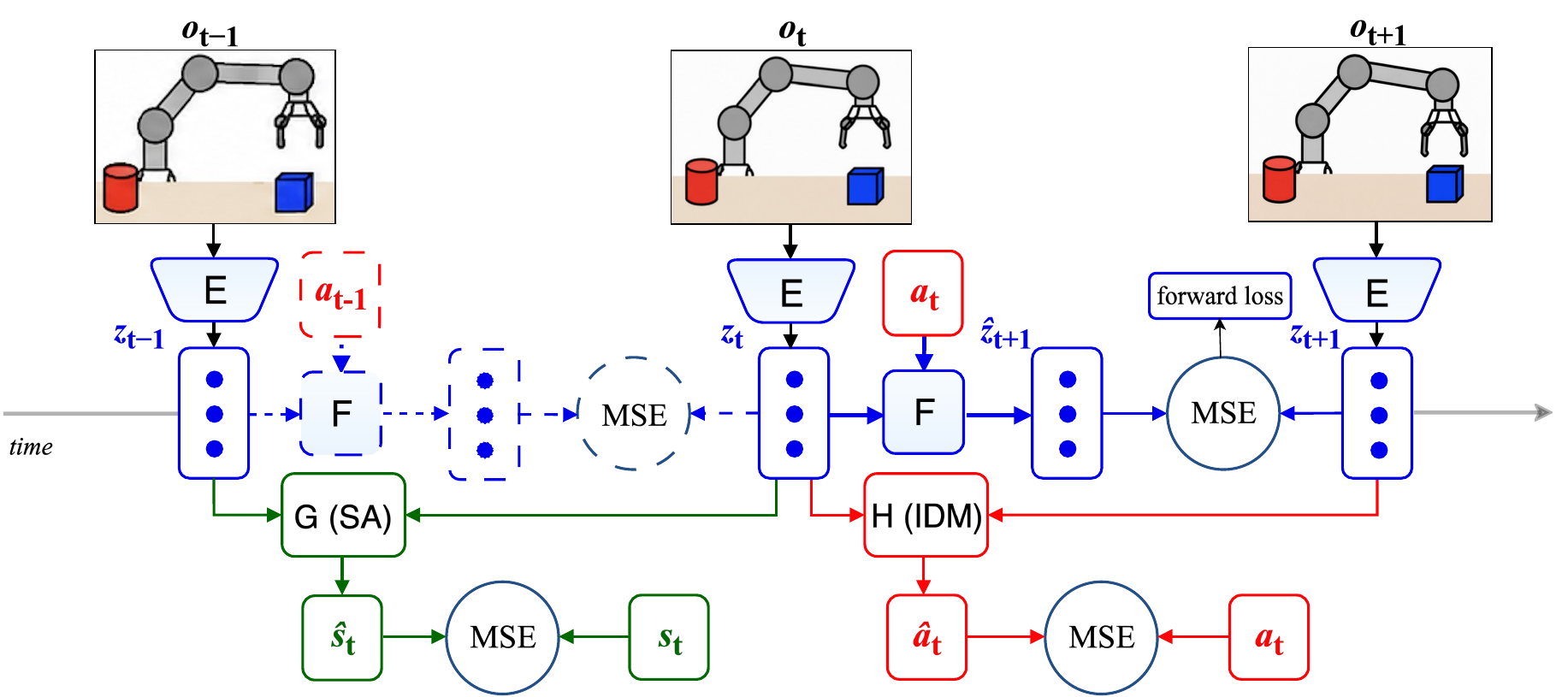}
  \caption{Overview of the proposed model. The encoder $E$ maps observations
  to latent representations, and the action-conditioned predictor $F$ models
  their evolution. The inverse-dynamics head $H$ predicts the executed action
  from consecutive representations, while the state-alignment
  head $G$ grounds them in measured physical states. The dashed branch denotes
  the corresponding prediction at the preceding time step.}
  \label{fig:method_overview}
\end{figure*}

\begin{figure*}[t]
  \centering
  \includegraphics[width=\textwidth]{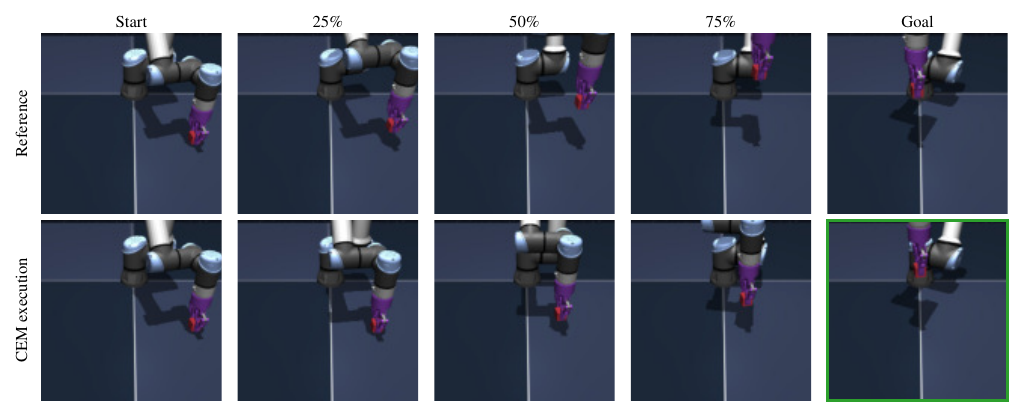}
  \caption{Qualitative goal-conditioned planning on OGBench-Cube. Top: the
  reference trajectory between the fixed start and goal observations. Bottom:
  actual simulator observations obtained by executing CEM actions.}
  \label{fig:cube_cem_rollout}
\end{figure*}

\section{Introduction}
\begingroup
\setlength{\parskip}{0pt}

World models have emerged as a promising paradigm for robotic autonomy by learning action-conditioned environment dynamics that support planning through imagined rollouts. Many visual world models~\cite{hafner2019planet,hafner2020dreamer} train latent dynamics with pixel-level observation prediction or reconstruction objectives, which can be computationally demanding and require modeling appearance variations, such as texture, illumination, and sensor noise, that are either irrelevant or even detrimental to downstream control. Action-conditioned Joint-Embedding Predictive Architectures (JEPAs)~\cite{lecun2022path} offer an appealing alternative by encoding raw observations into compact latent representations and predicting how these representations evolve under actions, without reconstructing pixels. Given a target representation encoded from a goal image, candidate action sequences are rolled out through the learned dynamics and evaluated against the target, enabling model-predictive control directly in latent space.

Learning a useful latent representation, however, remains a central challenge for JEPA-style world models, as the latent prediction objective alone admits a trivial solution that maps all observations to the same representation. Prior work addresses this issue using different learning strategies. DINO-WM~\cite{zhou2025dinowm} freezes a pretrained DINOv2 visual encoder~\cite{oquab2024dinov2} and learns action-conditioned dynamics over its patch-level features. This strategy avoids collapse without additional regularization, but relies on large-scale visual pretraining and computationally expensive patch-level rollouts. In contrast, LeWorldModel~\cite{maes2026leworldmodel} stabilizes end-to-end learning by matching latent embeddings to an isotropic Gaussian with SIGReg~\cite{balestriero2025lejepa}. While simple and effective, this distributional prior does not specify which aspects of the physical state should be retained in the representation for downstream control. PLDM~\cite{sobal2025rewardfree} learns representations end-to-end through a composite objective that combines latent prediction with VICReg-style variance--covariance regularization~\cite{bardes2022vicreg}, temporal smoothness, and inverse dynamics. In this formulation, inverse dynamics is one component of a broader regularization scheme. More recent approaches retain latent prediction while focusing the auxiliary objective on inverse dynamics: SMWM predicts the executed action from the concatenation of two consecutive latent representations~\cite{ivashkov2026sensorimotor}, whereas Delta-JEPA~\cite{zhang2026deltajepa} bases the prediction directly on their difference.

These action-recovery objectives make latent transitions informative of executed actions, but do not directly ground the same transitions in the associated physical configuration and motion. We investigate whether aligning latent representations with measured physical states can provide a complementary training signal for an end-to-end JEPA world model. Our method complements inverse dynamics (IDM) with a pair-based state-alignment (SA) objective: the former predicts executed actions, whereas the latter aligns consecutive latent representations with their corresponding measured physical states. We evaluate the resulting model on four goal-conditioned planning tasks and examine the temporal organization of its learned representations through straightening and transition-subspace analyses.

Our contributions are threefold:\par
\begingroup
\setlength{\parindent}{0pt}
\setlength{\parskip}{0pt}
\setlength{\leftskip}{1em}
\newcommand{\contributionitem}[2]{%
  \par\noindent\hangindent=2em\hangafter=1%
  \makebox[2em][l]{(#1)}#2}
\contributionitem{1}{We introduce an end-to-end JEPA world model that aligns
latent representations with measured physical states.}
\contributionitem{2}{Experiments across four benchmark tasks show that state
alignment consistently improves planning success over inverse dynamics alone.}
\contributionitem{3}{Transition-subspace analysis reveals that higher temporal
straightening, often regarded as desirable, can coincide with a substantially
lower effective transition dimension and weaker planning performance.}
\par
\endgroup
\endgroup

\section{Method}

\subsection{Offline Data}
We consider a fixed offline dataset collected by a behavior
policy. Let $\mathcal{D}=\{\tau^{(n)}\}_{n=1}^{N}$ denote the dataset. Each
trajectory is written as
$\tau=(o_0,s_0,a_0,o_1,s_1,\ldots,a_{T-1},o_T,s_T)$. Here, $o_t$ is a visual
observation at time $t$, $s_t$ denotes the corresponding physical measurements, and $a_t$ is an action chunk containing a fixed number of low-level actions executed between $o_t$ and $o_{t+1}$.

\subsection{Model Architecture}

Figure~\ref{fig:method_overview} summarizes our proposed model and its training
framework. The backbone of our method is an action-conditioned JEPA world
model~\cite{lecun2022path}, consisting of a visual encoder $E$ and a latent predictor $F$. The latent encoding and action-conditioned prediction are given by
\begin{align}
  z_t &= E(o_t),
  \label{eq:encoder}\\
  \hat z_{t+1} &= F\!\left(z_t,a_t\right),
  \label{eq:dynamics}
\end{align}
respectively. The directly encoded next-observation representation $z_{t+1}$ serves as the
target in the latent prediction objective,
\begin{equation}
  \mathcal{L}_{\mathrm{pred}}
  = \mathbb{E}_{(o_{t},a_{t},o_{t+1})\sim\mathcal{D}}
    \left[\left\|\hat z_{t+1}-z_{t+1}\right\|_2^2\right].
  \label{eq:basic_prediction}
\end{equation}
This objective is optimized end-to-end, updating the visual encoder (E) and the latent predictor (F).

The latent prediction loss avoids reconstructing future pixels, but by itself
admits a fully collapsed solution in which all observations are mapped to the
same representation. To discourage this solution, we adopt an inverse-dynamics
objective following prior
work~\cite{sobal2025rewardfree,ivashkov2026sensorimotor}:
\begin{equation}
  \mathcal{L}_{\mathrm{idm}}
    = \mathbb{E}_{(o_t,a_t,o_{t+1})\sim\mathcal{D}}
      \left[\|H(z_t,z_{t+1})-a_t\|_2^2\right],
  \label{eq:idm_expectation}
\end{equation}
where $H$ predicts the executed action from two consecutive representations.
This objective discourages full collapse because identical latent pairs across
transitions restrict $H$ to a constant-action predictor, whereas achieving an
error below this constant-predictor baseline requires action-predictive
variation. Gradients from $\mathcal{L}_{\mathrm{idm}}$ therefore encourage $E$
to distinguish latent pairs associated with different
actions~\cite{ivashkov2026sensorimotor}.

Inverse dynamics encourages latent pairs to be informative of executed
actions, but does not directly anchor them to observed physical measurements.
We therefore introduce a complementary state-alignment objective,
\begin{equation}
  \mathcal{L}_{\mathrm{sa}}
    = \mathbb{E}_{(o_{t-1},o_t,s_t)\sim\mathcal{D}}
      \left[\|G(z_{t-1},z_t)-s_t\|_2^2\right],
  \label{eq:state_expectation}
\end{equation}
where $s_t$ denotes the available physical
measurements. A consecutive latent pair is used because it provides the
temporal context needed to align velocity measurements in addition to the
instantaneous physical configuration. The full training objective is
\begin{equation}
  \mathcal{L}_{\mathrm{total}}
  = \mathcal{L}_{\mathrm{pred}}
  + \alpha\mathcal{L}_{\mathrm{sa}}
  + \beta\mathcal{L}_{\mathrm{idm}}.
  \label{eq:objective}
\end{equation}

Physical measurements provide training-only supervision; deployment performs
planning entirely in latent space using $E$ and $F$.

\subsection{Implementation}
To instantiate the JEPA backbone described above, we follow the lightweight architecture of LeWorldModel~\cite{maes2026leworldmodel}. The encoder is a randomly initialized
ViT-Tiny/14 whose \texttt{[CLS]} token is projected to a 192-dimensional
representation. The predictor is implemented as a six-layer causal Transformer
operating on a context of up to three recent latent--action pairs. For
stable regression, actions and Euclidean state quantities are standardized per
dimension using dataset statistics, while angular state quantities are expressed
in continuous coordinates rather than as raw angles to avoid wrap-around
discontinuities.
Each action chunk is then embedded into a
192-dimensional vector and incorporated at every predictor layer through
zero-initialized adaptive layer normalization (AdaLN)~\cite{maes2026leworldmodel}.
The state-alignment and inverse-dynamics heads, $G$ and $H$, are two-layer MLPs applied to
concatenated representation pairs, with hidden dimensions 256 and 512,
respectively. We optimize all components jointly using AdamW.

\subsection{Latent Planning}

At deployment, the encoder $E$ and predictor $F$ are held fixed. Given a
current observation $o_t$ and a goal image $o_g$, the encoder produces their
representations $z_t$ and $z_g$. For a candidate action sequence
$\mathbf{a}_{t:t+K-1}$, the predictor is rolled out recursively to obtain the
terminal representation $\hat z_{t+K}(\mathbf{a})$. Planning optimizes the
sequence according to the terminal latent distance
\begin{equation}
  \mathbf{a}^{*}_{t:t+K-1}
  = \arg\min_{\mathbf{a}_{t:t+K-1}}
    \left\|\hat z_{t+K}(\mathbf{a})-z_g\right\|_2^2.
  \label{eq:planning}
\end{equation}
We solve this optimization using the Cross-Entropy Method
(CEM)~\cite{rubinstein2004crossentropy}. CEM repeatedly samples candidate
action sequences, evaluates them through latent rollouts, and updates its
sampling distribution using the elite candidates. The optimized sequence is
then executed, after which planning is repeated from the updated observation
until the goal is reached or the interaction budget is exhausted.

\section{Experiments}

\subsection{Experimental Setup}

We adopt the four-task benchmark and evaluation protocol of
LeWorldModel~\cite{maes2026leworldmodel}: TwoRoom, Reacher, PushT, and
OGBench-Cube. These tasks cover visual navigation, articulated reaching,
planar object pushing, and robot-arm manipulation, respectively.
For state alignment, we use the full physical state provided by each task as
supervision.
Following this protocol, we evaluate each of our variants on the same 50 fixed
start--goal problems, with the goal 25 environment steps ahead and an
interaction budget of 50 steps.
Based on the planning success rate on the PushT validation split, we choose the
tied auxiliary-loss weights $\alpha=\beta=1.0$ from $\{0.01,0.1,1.0\}$ and use
this setting for all four tasks.

\subsection{Planning Performance in Latent Space}

We compare our method with DINO-WM~\cite{zhou2025dinowm},
PLDM~\cite{sobal2025rewardfree}, and
LeWorldModel~\cite{maes2026leworldmodel} in
Table~\ref{tab:planning_success}. Our method attains the highest success rates
on TwoRoom, PushT, and OGBench-Cube, while performing comparably to
LeWorldModel on Reacher. We further assess the contribution of state
alignment using the IDM-only ablation ($\alpha=0,\beta=1.0$) reported in
Table~\ref{tab:planning_success}. Adding state alignment consistently improves
planning success across all four tasks.

A successful OGBench-Cube trial provides a qualitative view of the resulting
control behavior in Fig.~\ref{fig:cube_cem_rollout}. Starting from the given
initial observation, the CEM-controlled simulator reaches the image-defined
goal within the evaluation budget.

\begin{table}[t]
    \caption{Goal-conditioned planning success rate (\%). Baseline mean values
    are taken from LeWorldModel~\cite{maes2026leworldmodel}; our results are
    averaged over three independently trained seeds, with 50 fixed planning
    problems evaluated per seed. Best results are shown in bold.}
    \label{tab:planning_success}
    \centering
    \setlength{\tabcolsep}{3.2pt}
    \begin{tabular}{lcccc}
      \toprule
      Method & TwoRoom & Reacher & PushT & OGB-Cube \\
      \midrule
      DINO-WM~\cite{zhou2025dinowm}
                   & \textbf{100} & 79 & 74 & 86 \\
      PLDM~\cite{sobal2025rewardfree}
                   & 97           & 78 & 78 & 65 \\
      LeWorldModel~\cite{maes2026leworldmodel}
                   & 87           & \textbf{86} & 96 & 74 \\
      \midrule

      IDM
                   & 94 & 63
                   & 83 & 85 \\
      SA+IDM (Ours)
                   & \textbf{100}
                   & 85
                   & \textbf{98}
                   & \textbf{87} \\
      \bottomrule
  \end{tabular}
\end{table}

\subsection{Temporal Straightening}

\begin{figure*}[t]
  \centering
  \includegraphics[width=\textwidth]{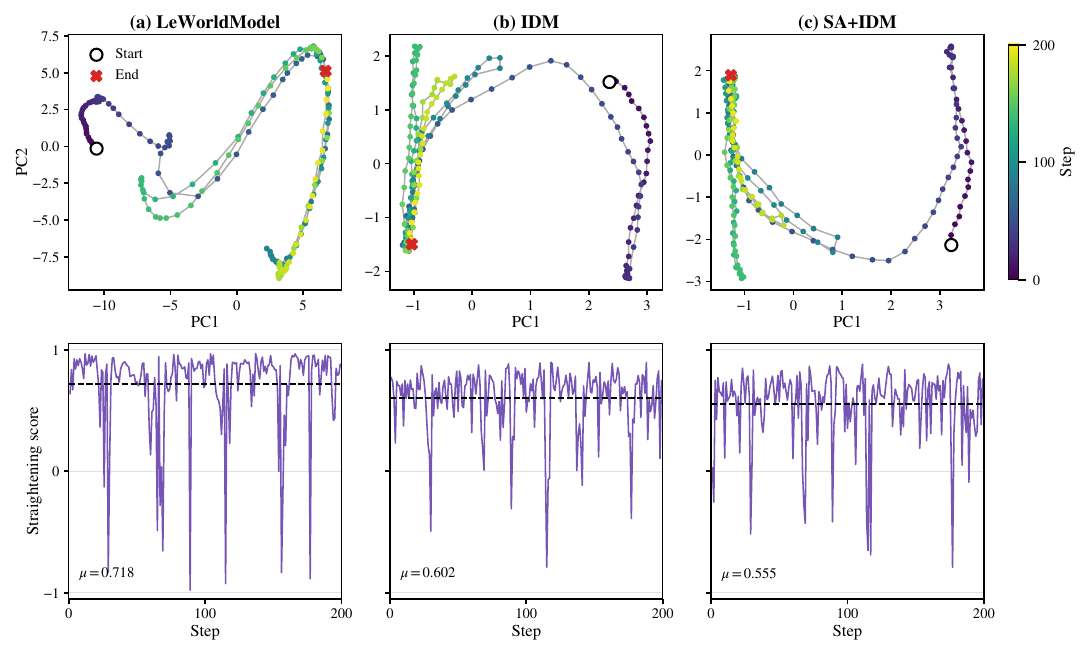}
  \caption{Temporal straightening on a representative OGBench-Cube trajectory.
  Top: separate PCA projections of the encoded latent states, colored by time;
  circles and crosses mark the start and end. Bottom: local straightening
  scores, with dashed lines indicating trajectory means. All scores are computed in the full 192-dimensional latent
  space. The same qualitative pattern was observed across all sampled
  trajectories.}
  \label{fig:cube_temporal_straightening}
\end{figure*}

Latent representations of temporally coherent physical trajectories are
generally expected to evolve smoothly with locally consistent
direction~\cite{henaff2019perceptual}. Temporal straightening quantifies this
property through the alignment of consecutive latent displacements. For a
latent trajectory $z_{1:T_i}^{(i)}$, with
$\Delta z_t^{(i)}=z_{t+1}^{(i)}-z_t^{(i)}$, the straightening score is defined
as follows~\cite{wang2026temporal}:

\begin{equation}
  S_{\mathrm{straight}}^{(i)}
  = \frac{1}{T_i-2}\sum_{t=1}^{T_i-2}
    \frac{\left\langle \Delta z_t^{(i)},\Delta z_{t+1}^{(i)}\right\rangle}
    {\left\|\Delta z_t^{(i)}\right\|_2
     \left\|\Delta z_{t+1}^{(i)}\right\|_2}.
  \label{eq:temporal_straightening}
\end{equation}

\begin{table}[t]
  \caption{Temporal straightening on OGBench-Cube. Values are mean $\pm$
  standard deviation over 100 fixed trajectories.}
  \label{tab:temporal_straightening}
  \centering
  \setlength{\tabcolsep}{6pt}
  \begin{tabular}{lc}
    \toprule
    Model & Straightening score \\
    \midrule
    LeWorldModel           & $0.69 \pm 0.025$ \\
    IDM                    & $0.62 \pm 0.029$ \\
    SA+IDM                 & $0.55 \pm 0.034$ \\
    \bottomrule
  \end{tabular}
\end{table}

Table~\ref{tab:temporal_straightening} reports the mean and standard deviation
of the straightening score over 100 sampled OGBench-Cube observation
trajectories held fixed across models.
The OGBench-Cube task is selected for this evaluation due to its rich 3D
translational and rotational variation.
We observe that adding state alignment to the IDM-only model lowers its mean
straightening score. This decrease may arise because explicit physical-state
supervision makes physical turns and curvature more strongly expressed in
latent displacements, thereby reducing the cosine alignment between successive
displacements.

LeWorldModel nevertheless attains the highest average straightening score
despite its lower planning success on OGBench-Cube. To inspect this discrepancy
at the trajectory level, Fig.~\ref{fig:cube_temporal_straightening} visualizes
the latent trajectories and local straightening scores for a representative
episode. The top row shows the temporal evolution of the encoded trajectories in two-dimensional PCA projections. Despite the arbitrary orientation of the independently fitted PCA bases, IDM and SA+IDM trace qualitatively similar global paths. Direction changes visible in the projections are reflected in decreases in the corresponding local straightening scores. LeWorldModel remains highly aligned for most steps but exhibits several isolated near-reversals toward $-1$, whereas IDM and SA+IDM exhibit more distributed and less extreme directional changes.

\begin{figure}[t]
  \centering
  \includegraphics[width=\columnwidth]{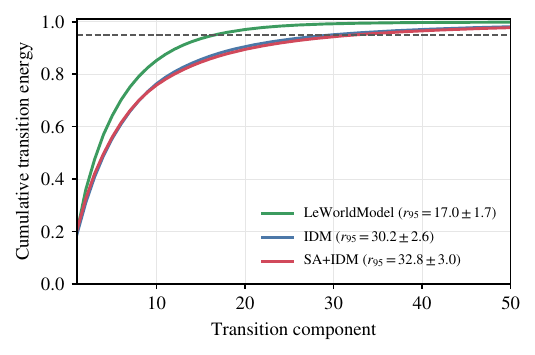}
  \caption{Transition-subspace analysis over 100 fixed OGBench-Cube
  trajectories. Curves show mean cumulative transition energy under an
  uncentered SVD of $\Delta z_t$. Here, $r_{95}$ is the number of components
  required to retain 95\% of the energy.}
  \label{fig:cube_transition_pca}
\end{figure}

We go one step further and use transition-subspace analysis to characterize the dominant
directions of temporal variation in latent space. For each of the same 100
trajectories, we apply an uncentered SVD to $\Delta z_t$ and compute $r_{95}$,
which is the number of components required to retain 95\% of the transition energy.
Fig.~\ref{fig:cube_transition_pca} reports the mean cumulative energy profiles
across trajectories. Across the 100 trajectories, the mean $r_{95}$ is $17$ for LeWorldModel,
compared with $30.2$ for IDM and $32.8$ for SA+IDM. LeWorldModel’s
higher straightening is therefore associated with a substantially
smaller effective transition dimension. This behavior is clearest in the rank-one limit, where successive nonzero displacements can only be parallel or antiparallel, so rare directional reversals produce a majority of
scores near $+1$ and a few near $-1$, yielding a high average with a polarized distribution. This argument is consistent with the polarized local straightening-score profile observed for LeWorldModel in Fig.~\ref{fig:cube_temporal_straightening}.

\section{Conclusion}

We presented a JEPA-style world model that combines inverse dynamics
with pair-based state alignment using available physical measurements. Across
four goal-conditioned tasks, state alignment consistently improves planning
success over the IDM-only ablation while remaining competitive with published
baselines.

Although our method exhibits lower average temporal straightening than
LeWorldModel, our transition-subspace analysis shows that LeWorldModel's higher straightening is associated with a substantially
lower effective transition dimension. This finding suggests that average
straightening alone can conceal the concentration of temporal variation in a low-dimensional subspace. SIGReg regularizes the embedding distribution across samples, whereas our analysis reveals that variation along trajectories can remain concentrated in a restricted transition subspace. Future work will investigate combining this distributional safeguard with physical grounding of latent transitions.

\section*{ACKNOWLEDGMENT}
The authors thank Ruiqing Cheng for insightful discussions.

\bibliographystyle{IEEEtran}
\bibliography{refs}

\end{document}